%% file: main.tex
\documentclass[10pt,twocolumn,letterpaper]{article}

\usepackage{cvpr}              

\input{preamble}

\definecolor{cvprblue}{rgb}{0.21,0.49,0.74}
\usepackage[pagebackref,breaklinks,colorlinks,allcolors=cvprblue]{hyperref}

\def\paperID{2} 
\def\confName{CVPR}
\def\confYear{2026}

\title{Not All NVFP4 QAT Recipes Are Equal: How Architecture and Scale Shape\\Model Quality for Anomaly Segmentation}

\author{
\begin{tabular}{c}
Zijian Du \quad Oleg Rybakov\\[2pt]
NVIDIA\\[2pt]
{\tt\small \{ldu, orybakov\}@nvidia.com}
\end{tabular}
}

\begin{document}
\maketitle
\input{sec/0_abstract}
\input{sec/1_intro}
\input{sec/2_related}
\input{sec/3_experimental_design}
\input{sec/4_experiments}
\input{sec/5_conclusion}


\end{document}

%% file: preamble.tex
\usepackage[accsupp]{axessibility}  
\usepackage{graphicx}
\usepackage{amsmath,amssymb}
\usepackage{booktabs}
\usepackage{placeins}
\usepackage{colortbl}
\usepackage{tikz}
\usetikzlibrary{arrows.meta,calc,positioning}

\usepackage{microtype}



%% file: sec/0_abstract.tex
\begin{abstract}
Real-time anomaly segmentation demands both high recall and efficient low-precision inference. We study the three-way interaction of model architecture, model scale, and FP4 quantization-aware training (QAT) recipe on a recall-critical brain tumor segmentation task, evaluating multiple architectures, scales, and QAT recipes under a unified protocol. We find that architecture choice has the largest impact on quantization robustness, with attention-based architectures showing remarkable resilience to recipe choice while CNN degrades under gradient-quantizing recipes at larger scales. At low capacity, FP4 can discretize softmax attention, but advanced QAT recipes prevent this collapse. At larger scales, advanced recipes mitigate gradient quantization noise that degrades CNN quality. Five-fold patient-level cross-validation confirms these findings are robust to data partition. Our results show that the Swin Transformer is robust to QAT recipe choice across all scales, making it the recommended architecture for FP4-quantized anomaly segmentation.
\end{abstract}

%% file: sec/1_intro.tex
\section{Introduction}
\label{sec:intro}

Pixel-level anomaly segmentation---detecting and localizing defects, lesions, or foreign objects in images---underpins critical applications from manufacturing inspection~\cite{chen2024surface,du2022lowdosage,du2023unsupervised} to medical diagnosis. The central difficulty is extreme class imbalance: anomalous regions are rare, and models that achieve high overall accuracy may still miss the very anomalies they are deployed to catch. In safety-critical domains, a missed defect (false negative) is far more costly than a false alarm, making recall the metric that matters most. Standard losses such as cross-entropy are ill-suited to this regime; focal loss~\cite{lin2017focal} down-weights easy negatives, while the Tversky loss~\cite{salehi2017tversky} provides direct control over the precision--recall trade-off by explicitly up-weighting false negatives. Evaluating such models also demands care: threshold-dependent metrics can be misleading under imbalance, motivating threshold-free alternatives such as the Area Under the Precision--Recall Curve (AUPRC)~\cite{saito2015auprc}.

FP4 quantization-aware training (QAT) offers the potential for significant inference speedup---up to $2\times$ throughput over FP8 on next-generation accelerators such as NVIDIA Blackwell~\cite{nvidia2025nvfp4,tseng2025mxfp4,xi2025fp4}. However, the critical question is not speed but \emph{quality}: can models trained at 4-bit precision retain the high recall of their full-precision baselines? For anomaly segmentation, any degradation in recall directly translates to missed defects or lesions in production.

The answer is not straightforward. Modern FP4 frameworks expose a menu of quantization recipes---combinations of block-scaling geometries, Random Hadamard Transforms (RHT), stochastic rounding (SR), and forward-only quantization---each introducing different noise profiles into the training dynamics~\cite{xi2025fp4}. At the same time, practitioners must choose among encoder--decoder CNNs~\cite{ronneberger2015unet}, Vision Transformers (ViT)~\cite{dosovitskiy2021vit}, and Swin Transformers~\cite{liu2021swin}---architectures that differ fundamentally in normalization (Batch Normalization vs.\ Layer Normalization), loss landscape geometry~\cite{bhattamishra2024lowsensitivity,hahn2024sensitivity,park2022vit}, and regularization requirements~\cite{touvron2021deit}. \textbf{It is therefore plausible that the best quantization recipe depends on the architecture, and that both choices interact with model scale---yet no prior work has studied these interactions jointly.}

We address this gap with a controlled cross-architecture study on brain tumor MRI segmentation~\cite{buda2019lgg}, a recall-critical anomaly detection task. We train a CNN, a Vision Transformer, and a Swin Transformer at three matched parameter scales under eight NVFP4 QAT recipes, using an identical pipeline with a recall-focused loss, threshold-free evaluation (AUPRC), and matched model sizes.\footnote{Code available at \url{https://github.com/ldu-nvidia/nvfp4-cvpr-2026}}

\begin{figure*}[t]
  \centering
  \begin{subfigure}{0.858\textwidth}
    \centering
    \includegraphics[width=\linewidth]{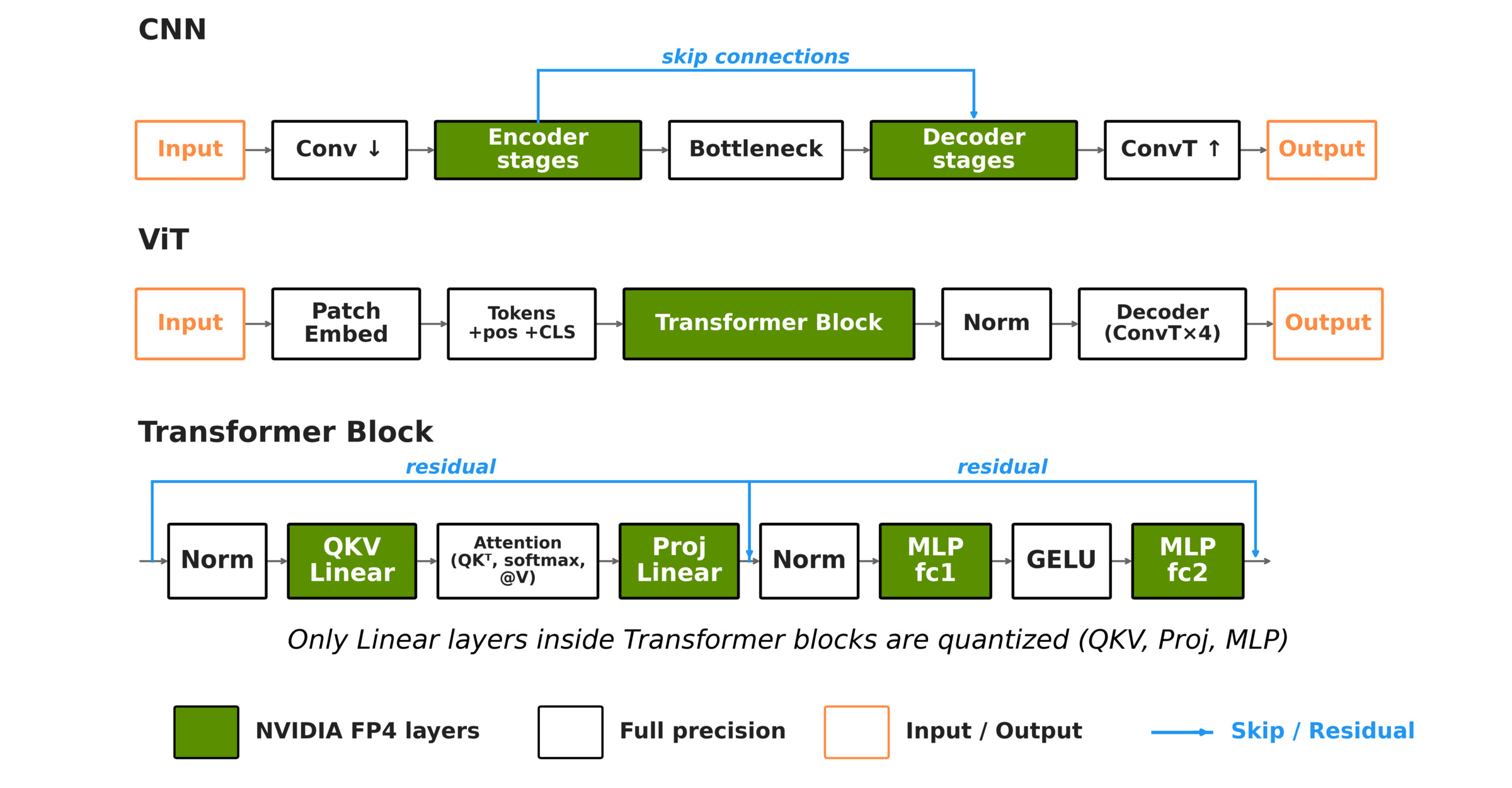}
    \caption{\textbf{CNN and ViT architectures} with quantized layers highlighted. Green = NVFP4-quantized; white = full precision.}
    \label{fig:arch_overview:arch}
  \end{subfigure}

  \vspace{6pt}
  \begin{subfigure}{0.288\textwidth}
    \centering
    \includegraphics[width=\linewidth]{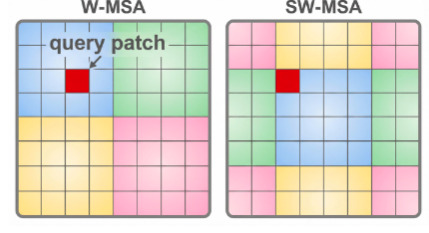}
  \end{subfigure}
  \hspace{0.02\textwidth}
  \begin{subfigure}{0.441\textwidth}
    \centering
    \includegraphics[width=\linewidth]{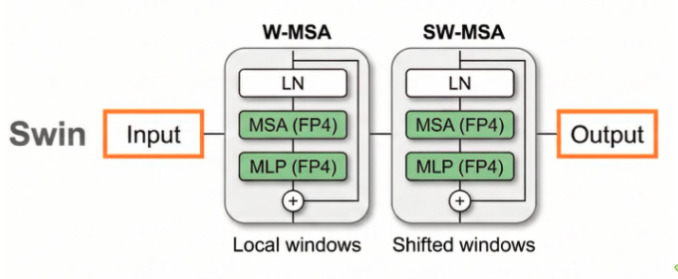}
  \end{subfigure}

  \vspace{4pt}
  \begin{subfigure}{0.98\textwidth}
    \caption{\textbf{Swin Transformer.} Windowed self-attention (W-MSA) restricts computation to local windows; shifted windows (SW-MSA) enable cross-window information flow. MSA and MLP layers are quantized to FP4.}
    \label{fig:arch_overview:swin}
  \end{subfigure}

  \caption{\textbf{Architecture overview and quantization layout.}}
  \label{fig:arch_overview}
\end{figure*}

Our main findings are:
\begin{enumerate}[leftmargin=*,topsep=2pt,itemsep=2pt]
  \item \textbf{Architecture choice determines FP4 quantization robustness.} Swin Transformer achieves the highest AUPRC at every matched parameter scale and is remarkably insensitive to recipe choice. CNN and ViT are more vulnerable, with ViT consistently underperforming regardless of optimizer. (\S\ref{sec:arch_compare}, \textbf{Figure~\ref{fig:eval_metrics}})
  \item \textbf{Advanced QAT recipes prevent quantization-induced softmax attention collapse.} At low capacity, FP4 discretizes attention weights into near-binary values, collapsing prediction quality. Advanced recipes resolve this failure mode entirely. (\S\ref{sec:small_fragile}, \textbf{Figure~\ref{fig:prc_dist}})
  \item \textbf{Advanced QAT recipes reduce gradient quantization error that degrades larger CNN models.} At scale, gradient quantization noise pushes CNN out of its optimum; advanced recipes---particularly stochastic rounding---recover the majority of the lost quality. (\S\ref{sec:quant_robustness}, \textbf{Figure~\ref{fig:qat_sensitivity}})
  \item \textbf{These findings are robust to data partition.} Five-fold patient-level cross-validation confirms these rankings hold consistently across independent patient splits.
\end{enumerate}

%% file: sec/2_related.tex
\section{Related Work}
\label{sec:related}
\paragraph{Class imbalance in anomaly segmentation.}
Anomalous regions typically occupy a small fraction of the image, biasing standard losses toward the majority class. The Tversky loss~\cite{salehi2017tversky,abraham2019focal,yeung2022unified} addresses this by providing direct control over the false-positive/false-negative trade-off, and AUPRC~\cite{saito2015auprc,mcdermott2024auprc} provides a threshold-free evaluation metric that is more discriminative than ROC-AUC under class imbalance. We adopt both in this work.
\paragraph{CNN vs.\ Vision Transformer for dense prediction.}
Encoder--decoder CNNs~\cite{ronneberger2015unet} benefit from strong spatial inductive biases, while Vision Transformers~\cite{dosovitskiy2021vit,ranftl2021dpt} capture global context through self-attention but require substantially more regularization~\cite{touvron2021deit}; transformers have since seen broad adoption in medical imaging~\cite{shamshad2023transformers}. Swin Transformers~\cite{liu2021swin} bridge the CNN--ViT gap with windowed attention that combines local and global context. Two properties of these architectures are particularly relevant to quantization: (1)~Batch Normalization in CNNs creates sharper loss landscapes and introduces training-time stochasticity~\cite{luo2019batchnorm,teye2018bayesian}, while attention-based architectures converge to flatter minima~\cite{park2022vit,bhattamishra2024lowsensitivity}---potentially affecting robustness to quantization noise; (2)~transformers exhibit lower input sensitivity than CNNs~\cite{hahn2024sensitivity}, suggesting they may tolerate coarser numerical representations. We test these hypotheses directly.
\paragraph{Low-precision training and FP4 quantization.}
Quantization-aware training (QAT) uses high-precision arithmetic to emulate low-precision numerics during training, allowing the model to learn to compensate for quantization noise before deployment. While 8-bit (FP8) QAT is well-established, 4-bit training has only recently become practical. Xi et al.~\cite{xi2025fp4} demonstrated fully quantized FP4 training for LLMs, using stochastic rounding in the backward pass and round-to-nearest in the forward pass. Tseng et al.~\cite{tseng2025mxfp4} showed that MXFP4 training with Random Hadamard Transforms achieves 1.3$\times$ speedup over FP8. NVIDIA's NVFP4 format~\cite{nvidia2025nvfp4} uses per-16-element FP8 scaling to provide finer granularity than MXFP4, and is natively supported on Blackwell GPUs with up to $2\times$ throughput over FP8. Edalati et al.~\cite{edalati2025bridging} compared MXFP4 and NVFP4, finding that NVFP4's smaller block size yields better accuracy. Lee et al.~\cite{lee2025tetrajet} identified weight oscillation as a primary accuracy degradation source in FP4 ViT training. Ozkara et al.~\cite{ozkara2025sr} provided theoretical grounding for stochastic rounding as implicit regularization under Adam-family optimizers. \textbf{Critically, nearly all FP4 work targets language models; controlled studies on vision architectures for dense prediction are absent.} We fill this gap by evaluating eight NVFP4 recipes across architectures and scales on a segmentation task.

%% file: sec/3_experimental_design.tex
\section{Experimental Design}
\label{sec:exp_design}

Our goal is to study how model architecture, model scale, and FP4 quantization recipe jointly affect segmentation quality. We vary these three dimensions while keeping the loss function, augmentation, and evaluation protocol fixed.

\subsection{Dataset}
\label{sec:dataset}
We use the LGG Brain MRI Segmentation dataset~\cite{buda2019lgg}, which contains 3,929 paired brain MRI slices and binary tumor masks from 110 patients. Positive (tumor-present) samples constitute 34.9\% of the data, representing moderate class imbalance typical of medical anomaly segmentation. We split the data \textbf{at the patient level} into 80/10/10 train/validation/test sets, ensuring that all slices from a given patient appear in exactly one partition and eliminating data leakage from correlated slices. We additionally validate our main findings with 5-fold patient-level cross-validation (\S\ref{sec:cv_robustness}).

\paragraph{Initialization.}
All recipes within a given \emph{(architecture, scale)} pair share the same initial weights, created by seeding PyTorch's random number generator with a fixed seed before constructing the model. Because different architectures have different parameter shapes, the resulting weight tensors are not identical across architectures---``same seed'' means the same RNG state, not the same numerical weights. For the QAT sensitivity analysis (\S\ref{sec:quant_robustness}), we additionally train with 10 seeds to establish statistical significance.

\subsection{Architectures}
\label{sec:architectures}
We compare three architectures at three matched parameter scales:

\paragraph{CNN.}
A scalable encoder--decoder CNN~\cite{ronneberger2015unet} with configurable depth and width. Each encoder stage applies a stride-2 convolution with Batch Normalization and ReLU. The decoder mirrors the encoder with transposed convolutions, and skip connections concatenate encoder features at each resolution. Channel widths are constrained to multiples of 16 for compatibility with FP4 block quantization. The first encoder layer, bottleneck layer, and last decoder layer are kept in full precision to preserve input fidelity and output quality.
\paragraph{Vision Transformer (ViT).}
A patch-based transformer~\cite{dosovitskiy2021vit} with $16\times 16$ patches, learned positional embeddings, and a decoder head that maps per-patch embeddings back to pixel-level predictions via reshaping. Each transformer block uses Layer Normalization, multi-head self-attention, and an MLP with GELU activation. We train ViT with both AdamW and Adamax optimizers to test whether optimizer choice affects quantization sensitivity; only Linear layers (QKV, projection, MLP) are quantized.
\paragraph{Swin Transformer.}
A shifted-window transformer~\cite{liu2021swin} with $16\times 16$ patch embedding and windowed multi-head self-attention. Consecutive blocks alternate between regular window attention (W-MSA) and shifted-window attention (SW-MSA), enabling cross-window information flow without the quadratic cost of global self-attention. Each block uses Layer Normalization, a pre-norm residual structure, and an MLP with GELU activation. A transposed-convolution decoder maps patch embeddings back to pixel-level predictions.
\paragraph{Matched scales.}
We design three scale tiers where all architectures have approximately equal parameter counts (\textbf{Table~\ref{tab:scales}}). This ensures that performance differences are attributable to architecture, not capacity. \textbf{Figure~\ref{fig:arch_overview}} illustrates the architectures and highlights which layers are quantized.

\begin{table}[t]
  \centering
  \caption{\textbf{Architecture configurations at matched parameter scales.}}
  \label{tab:scales}
  \footnotesize
  \setlength{\tabcolsep}{1.5pt}
  \begin{tabular}{@{}llll@{}}
    \toprule
    \textbf{Scale} & \textbf{CNN} & \textbf{ViT} & \textbf{Swin} \\
    \midrule
    ${\sim}$530K & 4 stg, 64ch & $d{=}64,L{=}10,h{=}2$ & $d{=}64,L{=}10,h{=}4,w{=}4$ \\
    ${\sim}$3.7M & 6 stg, 128ch & $d{=}192,L{=}9,h{=}6$ & $d{=}192,L{=}9,h{=}6,w{=}8$ \\
    ${\sim}$13.7M & 7 stg, 224ch & $d{=}448,L{=}8,h{=}14$ & $d{=}448,L{=}8,h{=}14,w{=}16$ \\
    \bottomrule
  \end{tabular}
\end{table}

\subsection{NVFP4 format}
\label{sec:nvfp4_format}
NVFP4~\cite{nvidia2025nvfp4,nvidia2025nvfp4train} represents each value in 4-bit floating point (E2M1: 1 sign, 2 exponent, 1 mantissa bit; discrete values from $-6$ to $6$). A full-precision tensor $X$ is quantized via two-level scaling:
\begin{flalign}
&\textbf{L1 (per-tensor):}~ S_t = \max|X|,\; X_1 = X / S_t & \nonumber\\
&\textbf{L2 (per-block):}~ S_b = \max|X_{1,b}|,\; q_b = \mathrm{FP4}(X_1 / S_b) & \nonumber
\end{flalign}
where $X_{1,b}$ is a contiguous 16-element micro-block. $S_t$ (FP32) maps the tensor range globally; $S_b$ (FP8 E4M3) provides fine-grained per-block adaptation. At dequantization: $\hat{X}_b \approx \mathrm{BF16}(q_b) \cdot S_b \cdot S_t$. This two-level design provides finer granularity than MXFP4's single power-of-two scale.

\subsection{NVFP4 QAT recipes}
\label{sec:nvfp4}
During QAT, each quantized linear layer emulates NVFP4 precision in high-precision arithmetic (\textbf{Figure~\ref{fig:qat_flow}}). The base recipe (NVFP4 Full) quantizes activations and weights in the forward pass and quantizes the upstream gradient in the backward pass. Three advanced techniques can be layered on top to reduce quantization error:

\begin{figure*}[t]
  \centering
  \includegraphics[width=0.75\textwidth]{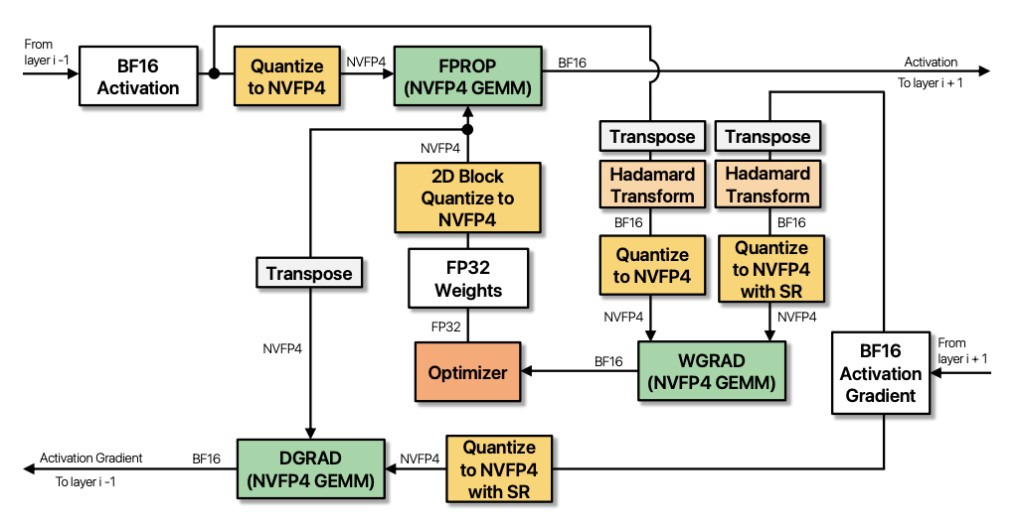}
  \vspace{2pt}
  \caption{\textbf{NVFP4 QAT data flow for a single quantized linear layer.} Yellow blocks: quantization steps; green blocks: NVFP4 GEMM operations; orange: optimizer. Each QAT recipe (\textbf{Table~\ref{tab:recipes}}) selectively enables or disables these steps.}
  \label{fig:qat_flow}
  \vspace{6pt}
\end{figure*}

\paragraph{Advanced techniques.}
\begin{itemize}[leftmargin=*,topsep=2pt,itemsep=2pt]
  \item \textbf{2D block-scaling.} Replaces $1\times 16$ (1D) blocks with $16\times 16$ blocks for weight quantization, making the quantized weight representation transpose-invariant between the forward and backward passes so that $Q(\mathbf{W})$ and $Q(\mathbf{W}^\top)$ use the same block boundaries.
  \item \textbf{Random Hadamard Transform (RHT).} Applies a randomized orthogonal rotation to activations and gradients before quantization in the WGRAD path, redistributing block-level outlier energy so that no single large value dominates the block scale.
  \item \textbf{Stochastic rounding (SR).} Replaces deterministic round-to-nearest on the upstream gradient with probabilistic rounding, eliminating the systematic bias that causes small gradient updates to be rounded to zero~\cite{ozkara2025sr}.
\end{itemize}

\paragraph{Eight recipes.}
We evaluate eight recipes (\textbf{Table~\ref{tab:recipes}}) that span the design space along four axes: (1)~whether 2D scaling and RHT are enabled, (2)~whether SR is applied to gradients, (3)~whether the backward pass uses quantized or full-precision tensors (forward-only), and (4)~whether the backward pass reuses the forward quantized values (Chain Rule) or re-quantizes independently. The \textbf{Baseline} trains entirely in BF16. \textbf{NVFP4 Full} applies base quantization to all tensors in both passes without any advanced technique. \textbf{Fwd-Only} and \textbf{Fwd+RHT} quantize only the forward pass, keeping backward in BF16. \textbf{Chain Rule} reuses the same $Q(\mathbf{X})$, $Q(\mathbf{W})$ from the forward pass in the backward pass, so gradients reflect the actual quantization noise. \textbf{SR Only} adds stochastic rounding to gradients. \textbf{2D+RHT} enables 2D scaling and RHT. \textbf{2D+RHT+SR} is the most advanced recipe, combining all three techniques.

\paragraph{QAT linear layer equations.}
\textbf{Figure~\ref{fig:qat_flow}} illustrates how the forward and backward passes are executed for a quantized linear layer. Equations~\ref{eq:fprop}--\ref{eq:wgrad} formalize this data flow for the most advanced recipe (2D+RHT+SR); all other recipes are variants that disable one or more steps (see \textbf{Table~\ref{tab:recipes}}):
\begin{flalign}
  &\textbf{Fwd:}~~ \mathbf{Y}^{\text{BF16}} = \hat{\mathbf{X}}^{\text{FP4}} \!\cdot\! \hat{\mathbf{W}}_{\text{2D}}^{\text{FP4}} \label{eq:fprop}&\\[3pt]
  &\textbf{DG:}~~ \tfrac{\partial L}{\partial \mathbf{X}}^{\text{BF16}} = Q_{\text{SR}}(\mathbf{G})^{\text{FP4}} \!\cdot\! \hat{\mathbf{W}}_{\text{2D}}^{\text{FP4}\top} \label{eq:dgrad}&\\[3pt]
  &\textbf{WG:}~~ \tfrac{\partial L}{\partial \mathbf{W}}^{\text{BF16}} = Q_{\text{SR}}(\text{RHT}(\mathbf{G}^{\!\top}))^{\text{FP4}} \!\cdot\! Q(\text{RHT}(\mathbf{X}^{\!\top}))^{\text{FP4}} \label{eq:wgrad}&
\end{flalign}
where $\hat{\mathbf{X}}$ is the quantized activation, $\hat{\mathbf{W}}_{\text{2D}}$ is the weight quantized with 2D block-scaling, $\mathbf{G}$ is the upstream BF16 gradient, $Q$ denotes NVFP4 quantization, and $Q_{\text{SR}}$ denotes quantization with stochastic rounding.

\vspace{2pt}
\begin{table*}[t]
  \centering
  \caption{\textbf{NVFP4 QAT recipes.} Each recipe selectively enables quantization techniques. ``2D'' applies 2D block-scaling to weights; ``RHT'' applies Random Hadamard Transform in the WGRAD path; ``SR'' applies stochastic rounding to the upstream gradient. ``Bwd W/X'' indicates whether backward-pass gradients use quantized (FP4) or original (BF16) tensors.}
  \label{tab:recipes}
  \small
  \setlength{\tabcolsep}{6pt}
  \begin{tabular}{l c c c c c c}
    \toprule
    \textbf{Recipe} & \textbf{2D} & \textbf{RHT} & \textbf{SR} & \textbf{Fwd-only} & \textbf{Bwd W/X} & \textbf{Upstream Gradient} \\
    \midrule
    Baseline (BF16) & -- & -- & -- & -- & -- & -- \\
    NVFP4 Full &  &  &  &  & FP4 & FP4 \\
    Fwd-Only &  &  &  & \checkmark & BF16 & BF16 \\
    Fwd+RHT &  & \checkmark &  & \checkmark & BF16 & BF16 \\
    Chain Rule$^\ast$ &  &  &  &  & FP4$^\ast$ & BF16 \\
    SR Only &  &  & \checkmark &  & FP4 & FP4 \\
    2D+RHT & \checkmark & \checkmark &  &  & FP4 & FP4 \\
    2D+RHT+SR & \checkmark & \checkmark & \checkmark &  & FP4 & FP4 \\
    \bottomrule
  \end{tabular}

  \vspace{2pt}
  \footnotesize{$^\ast$Chain Rule reuses the same quantized $Q(\mathbf{W})$, $Q(\mathbf{X})$ from the forward pass in the backward pass, rather than re-quantizing or using BF16 originals.}
  \vspace{4pt}
\end{table*}

\subsection{Training Protocol and Evaluation}
\label{sec:setup}
All architectures share an identical training pipeline except for the optimizer, which is matched to each architecture family:

\begin{table}[t]
  \centering
  \caption{\textbf{Optimizer configurations per architecture.}}
  \label{tab:optimizers}
  \footnotesize
  \setlength{\tabcolsep}{3pt}
  \begin{tabular}{@{}llccc@{}}
    \toprule
    \textbf{Architecture} & \textbf{Optimizer} & \textbf{LR} & \textbf{WD} & \textbf{Purpose} \\
    \midrule
    CNN          & Adamax & $5\!\times\!10^{-4}$ & 0 & CNN baseline \\
    ViT-Adamax   & Adamax & $5\!\times\!10^{-4}$ & 0 & Optimizer probe \\
    ViT-AdamW    & AdamW  & $1\!\times\!10^{-4}$ & $10^{-4}$ & Transformer-optimal \\
    Swin         & AdamW  & $1\!\times\!10^{-4}$ & $10^{-4}$ & Transformer-optimal \\
    \bottomrule
  \end{tabular}
\end{table}
\vspace{-4pt}
\begin{itemize}[leftmargin=*,topsep=2pt,itemsep=2pt]
  \item \textbf{Training loss:} BCE + Tversky, a recall-focused objective that penalizes false negatives $2.3\times$ more than false positives:
  \begin{equation}
    \label{eq:loss}
    L_{\mathrm{train}} \!=\! \mathrm{BCE}(\hat{y},y) + 1 \!-\! \frac{\mathrm{TP}_s}{\mathrm{TP}_s + \alpha\,\mathrm{FP}_s + \beta\,\mathrm{FN}_s}
  \end{equation}
  where $\hat{y}\!=\!\sigma(z)$, \; $\mathrm{TP}_s\!=\!\sum\hat{y}\,y$, \; $\mathrm{FP}_s\!=\!\sum\hat{y}(1\!-\!y)$, \; $\mathrm{FN}_s\!=\!\sum(1\!-\!\hat{y})y$, \; and $\alpha\!=\!0.3$, $\beta\!=\!0.7$.

  \item \textbf{Scheduler:} ReduceLROnPlateau (factor 0.5, patience 10).
  \item \textbf{Early stopping:} Monitor validation loss, patience 20, warmup 40 epochs.
  \item \textbf{Augmentation (training only):} Resize to 256$\times$256, horizontal/vertical flip, brightness--contrast, shift--scale--rotate, ImageNet normalization. Validation and test sets are only resized and normalized.
  \item \textbf{Initialization:} All recipes within a model size share the same fixed random seed for fair comparison.

  \item \textbf{Evaluation metric:} Area Under the Precision--Recall Curve (AUPRC), a threshold-free metric computed on the held-out test set using the best checkpoint selected by validation loss. Unlike ROC-AUC, AUPRC does not benefit from the large true-negative count under class imbalance~\cite{saito2015auprc}, making it more discriminative for our setting.
\end{itemize}

%% file: sec/4_experiments.tex
\section{Experiments}
\label{sec:experiments}

\begin{figure*}[t]
  \centering
  \includegraphics[width=0.85\textwidth,trim={0.2cm 0.2cm 0.2cm 0.2cm},clip]{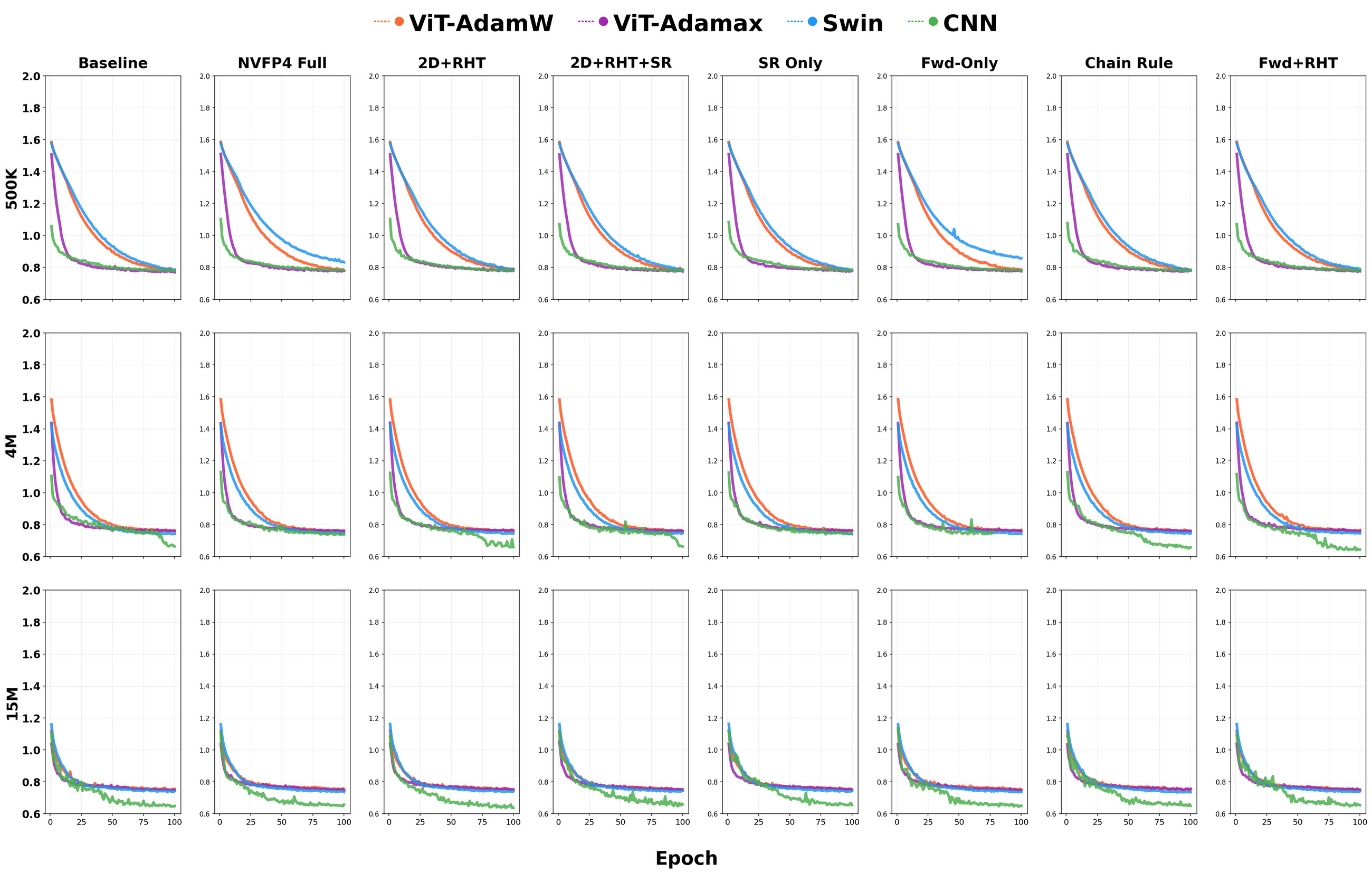}
  \caption{\textbf{Validation loss curves} (BCE + Tversky) for all architectures across 8 NVFP4 recipes (columns) and 3 model scales (rows: 500K, 4M, 15M). All architectures converge across recipes.}
  \label{fig:val_loss}
\end{figure*}

Data is split at the patient level (\S\ref{sec:dataset}) so that no patient's slices appear in more than one partition. To further isolate recipe effects from initialization variance, we repeat the QAT sensitivity analysis (\S\ref{sec:quant_robustness}) with 10 independent random seeds and report mean $\pm$ 95\% confidence interval; we additionally confirm the main findings with 5-fold patient-level cross-validation (\S\ref{sec:cv_robustness}). \textbf{Figure~\ref{fig:val_loss}} shows the validation loss curves across all configurations.

\begin{figure}[t]
  \centering
  \includegraphics[width=\columnwidth]{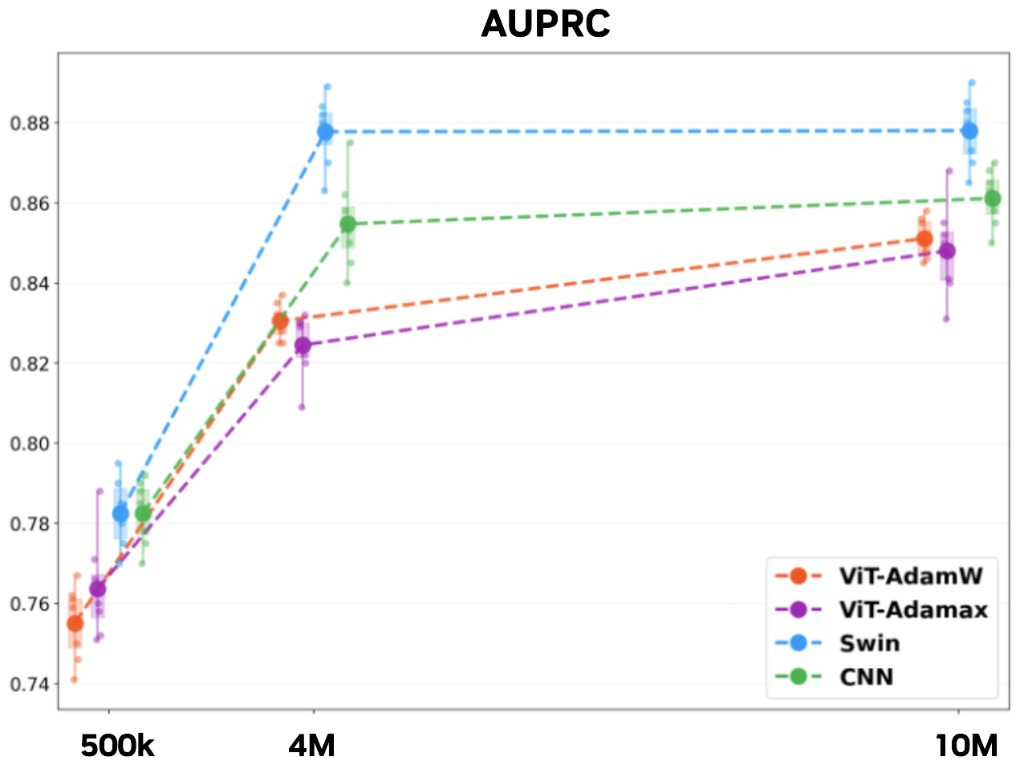}
  \caption{\textbf{AUPRC vs.\ model scale for all architectures.} AUPRC increases with scale but with diminishing returns from 4M to 15M. Swin achieves the highest AUPRC at every scale.}
  \label{fig:eval_metrics}
\end{figure}

\subsection{Model scale}
\label{sec:scale_dominant}
Across all architectures and recipes, increasing model size from 500K to 4M yields the largest AUPRC gains (\textbf{Figure~\ref{fig:eval_metrics}}). All architectures show substantial improvement in this range. However, the gain from 4M to 15M is marginal for all architectures. This diminishing return likely reflects overfitting on the relatively small MRI dataset (${\sim}$3,000 training images): the 15M models have sufficient capacity to memorize the training set, but the additional parameters provide little additional generalization. This supports 4M as the practical sweet spot---it captures most of the capacity benefit without the overfitting risk.

\subsection{Model architecture}
\label{sec:arch_compare}
At every scale, the Swin Transformer achieves competitive AUPRC (\textbf{Figure~\ref{fig:eval_metrics}}), with statistically significant advantages over CNN at 4M ($p=0.0012$) and 15M ($p=0.0208$). The ViT consistently underperforms both CNN and Swin, regardless of whether AdamW or Adamax is used as the optimizer---ruling out optimizer choice as the explanation for ViT's lower performance.

We attribute Swin's advantage to windowed attention's \emph{local-global} inductive bias~\cite{liu2021swin}. Like convolution, windowed attention restricts each layer's receptive field to a local neighborhood, exploiting the spatial locality inherent in images. Unlike convolution, shifting the window partition every other layer enables information flow across boundaries, gradually building global context without the quadratic cost of full self-attention. Standard ViT lacks this locality prior: it attends over all patches from the first layer, which is data-hungry and less effective on small medical datasets where spatial correlations are strong~\cite{dosovitskiy2021vit,touvron2021deit}.

\begin{figure*}[t!]
  \centering
  \includegraphics[width=0.80\textwidth]{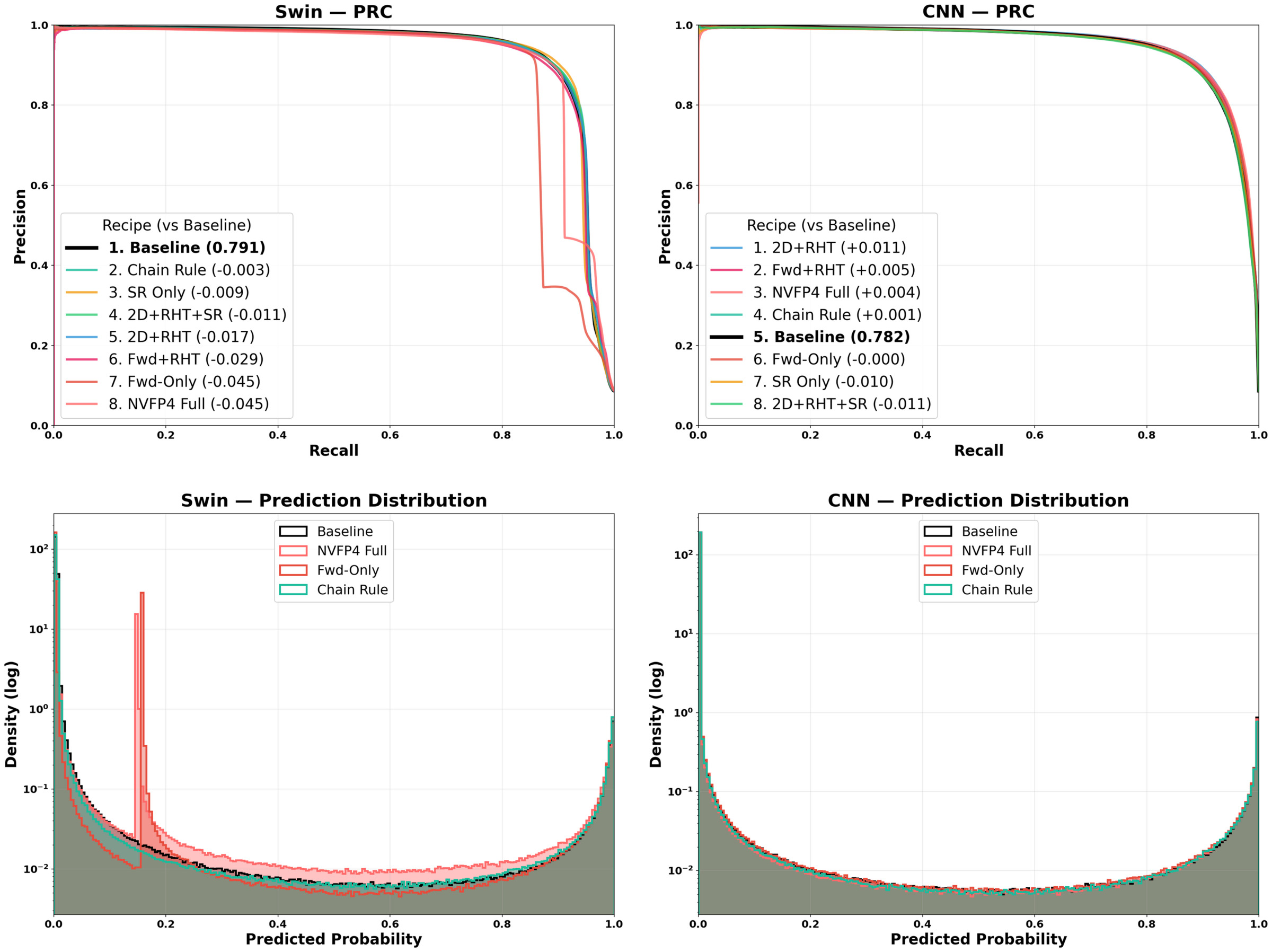}
  \caption{\textbf{PRC curves and prediction distributions at 500K scale.} Swin (left) shows discretized prediction probabilities under NVFP4 Full and Forward-Only, while CNN (right) remains smooth, resulting in lower AUPRC for Swin at this scale.}
  \label{fig:prc_dist}
\end{figure*}

\subsection{Advanced recipes prevent attention discretization in small Swin}
\label{sec:small_fragile}
At 500K scale, Swin's AUPRC drops below CNN's under certain recipes---an unexpected reversal given Swin's advantage at larger scales. To understand this, we examined the PRC curves and prediction distributions at 500K, which revealed a recipe-dependent discretization effect unique to attention-based architectures.

\paragraph{Why small Swin is vulnerable.}
At 500K, Swin's head dimension is only $d_h\!=\!16$. Each attention logit is a dot product of 16 FP4 values, giving an SNR $\approx 4$. Softmax exponentiates these coarse logits ($\alpha_j \propto e^{\ell_j}$), amplifying small quantization errors into near-binary attention weights that discretize predictions~\cite{hahn2024sensitivity}. CNN is immune: its convolution--BN--ReLU path is Lipschitz-continuous~\cite{luo2019batchnorm} and cannot amplify noise. \textbf{Figure~\ref{fig:prc_dist}} confirms this---Swin's PRC curves develop staircase artifacts and its histograms show discrete spikes under the affected recipes, while CNN remains smooth throughout.

\paragraph{RHT and SR prevent the collapse.}
The two affected recipes---NVFP4 Full and Forward-Only---are the only quantized recipes (besides Chain Rule) lacking \emph{both} RHT and SR (\textbf{Table~\ref{tab:recipes}}). RHT directly smooths the forward-pass logits by redistributing outlier energy before quantization; SR provides unbiased gradient rounding so QAT can learn $Q,K$ projections that keep logits well-separated. All RHT- or SR-equipped recipes stay within 0.017 of baseline at 500K. This demonstrates a concrete benefit of these techniques beyond general accuracy: \textbf{they make low-capacity transformers robust to the attention-discretization failure mode that bare FP4 quantization introduces}.

\paragraph{Forward--backward consistency is sufficient without advanced techniques.}
Chain Rule also lacks RHT and SR yet loses only 0.003 AUPRC, because it reuses the \emph{same} quantized values in both forward and backward passes. This forward--backward consistency lets gradients reflect the actual quantization noise, enabling QAT to compensate. NVFP4 Full breaks this by independently quantizing the gradients (too noisy at low capacity); Forward-Only breaks it by using BF16 originals in the backward pass (gradients blind to quantization). At 4M+ ($d_h \!\geq\! 32$), logit SNR is sufficient for all recipes and the distinction vanishes.

\begin{figure*}[t!]
  \centering
  \includegraphics[width=0.88\textwidth]{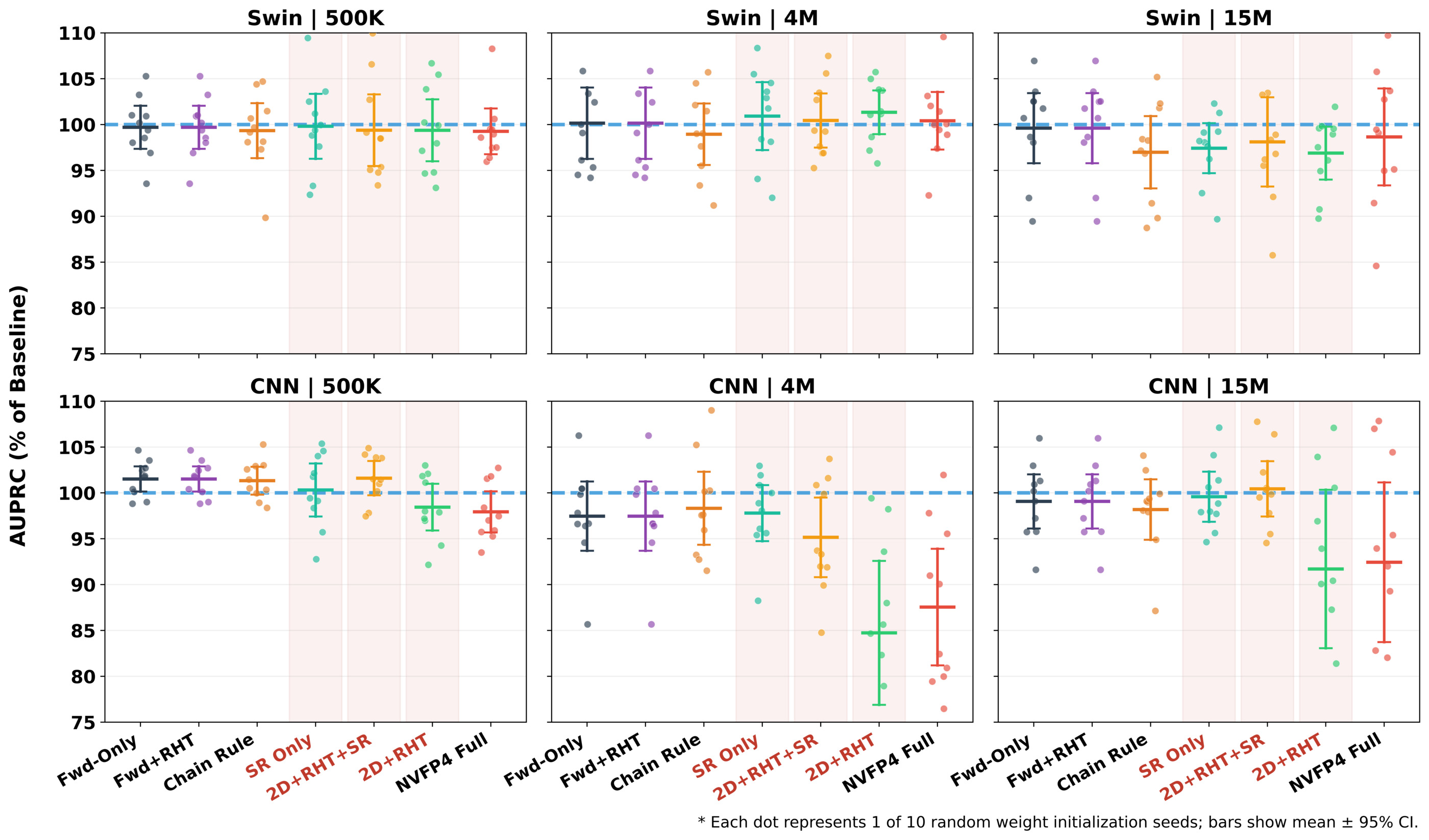}
  \caption{\textbf{Normalized AUPRC (\% of BF16 baseline) across all quantized recipes} for Swin (top) and CNN (bottom) at three scales. Each dot is one of 10 random weight initialization seeds; bars show mean $\pm$ 95\% CI. Shaded region highlights recipes incorporating SR, RHT, or 2D scaling.}
  \label{fig:qat_sensitivity}
\end{figure*}

\subsection{Advanced recipes mitigate gradient noise in large CNN}
\label{sec:quant_robustness}
Using 10 independent random seeds per configuration, we normalize each run's AUPRC to the BF16 baseline to isolate recipe effects (\textbf{Figure~\ref{fig:qat_sensitivity}}). Across all three scales, Swin's AUPRC stays within 97--103\% of baseline regardless of recipe. CNN, however, degrades sharply at 4M and 15M under gradient-quantizing recipes---particularly NVFP4 Full and 2D+RHT, which drop to ${\sim}84\%$ and ${\sim}92\%$ of baseline respectively. Among the advanced techniques (highlighted in \textbf{Figure~\ref{fig:qat_sensitivity}}), stochastic rounding contributes the most to CNN quality recovery: comparing SR~Only against NVFP4~Full isolates SR's effect, showing it recovers the majority of the lost AUPRC. RHT and 2D scaling provide additional but smaller gains on top of SR, and the full 2D+RHT+SR recipe brings CNN closest to baseline.

This pattern is consistent with the hypothesis that attention-based architectures converge to flatter minima~\cite{park2022vit,bhattamishra2024lowsensitivity}, where FP4 gradient noise stays within the basin of the optimum, while batch-normalized CNNs occupy sharper basins~\cite{luo2019batchnorm} where quantized gradients can push parameters out of the minimum. Recent FP4 training studies corroborate this, reporting gradient sensitivity in CNNs and recommending SR or RHT as mitigations~\cite{xi2025fp4,tseng2025mxfp4,lee2025tetrajet}.

\subsection{Cross-validation robustness}
\label{sec:cv_robustness}
\begin{figure}[t]
  \centering
  \includegraphics[width=0.95\columnwidth]{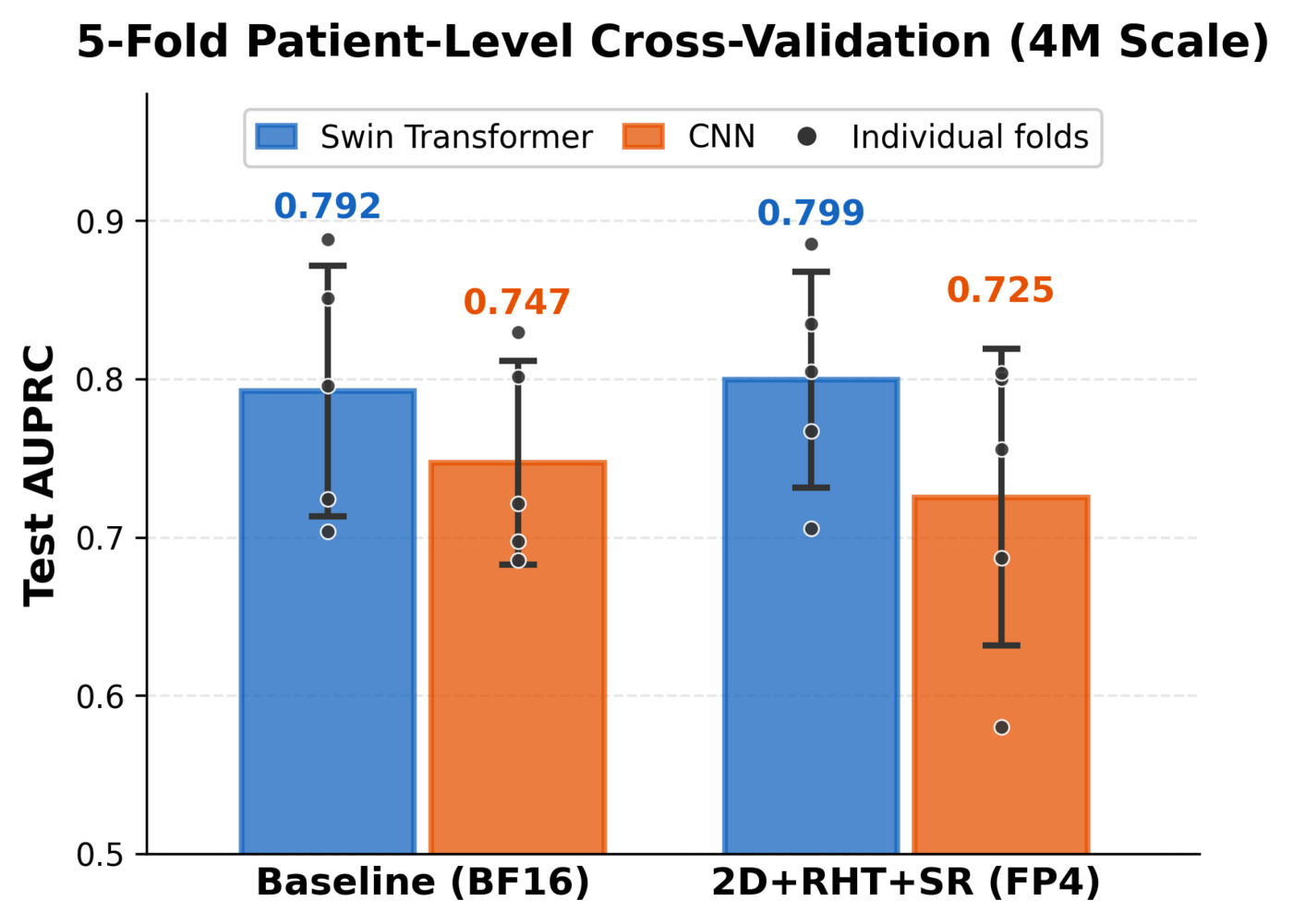}
  \caption{\textbf{Cross-validation robustness.} Swin outperforms CNN under both BF16 baseline and the best FP4 recipe across all five patient-level folds at 4M scale. Dots show individual folds; error bars show $\pm$1 std.}
  \label{fig:cv_robustness}
\end{figure}

To verify that our findings are not artifacts of a single data partition, we retrain the 4M Swin and CNN under both BF16 baseline and the recommended 2D+RHT+SR recipe using 5-fold patient-level cross-validation, where each fold's train, validation, and test sets contain non-overlapping patients (\textbf{Figure~\ref{fig:cv_robustness}}). Swin achieves a mean test AUPRC of $0.792\pm0.079$ under baseline and $0.799\pm0.068$ under FP4---virtually identical, confirming its recipe robustness. CNN scores $0.747\pm0.065$ (baseline) and $0.725\pm0.094$ (FP4), consistently below Swin. The Swin--CNN gap ($\Delta{\approx}0.05$--$0.07$) is stable across all five folds, confirming that the architectural ranking reported in \S\ref{sec:arch_compare} is robust to data split variation.

%% file: sec/5_conclusion.tex
\section{Conclusion}
\label{sec:conclusion}
We studied NVFP4 QAT across three architectures, three matched scales, and eight recipes on recall-critical brain tumor segmentation.
\begin{enumerate}[leftmargin=*,topsep=1pt,itemsep=1pt]
\item \textbf{Swin is the most FP4-robust architecture}, achieving the highest AUPRC at every scale and remaining insensitive to recipe choice. CNN degrades under gradient-quantizing recipes; ViT underperforms regardless of optimizer. Cross-validation confirms this ranking.
\item \textbf{Advanced QAT recipes are essential}, resolving two distinct failure modes: attention discretization in small transformers and gradient noise in large CNNs.
\item \textbf{4M is the practical sweet spot}; the 4M Swin with 2D+RHT+SR is the recommended configuration. Knowledge distillation and pruning could further reduce size.
\end{enumerate}

\paragraph{Limitations.}
We use a single dataset; generalization remains future work. We do not measure FP4 inference latency. Mechanistic explanations are literature-consistent but not Hessian-verified. Interaction with ImageNet pretraining is unexplored.